\documentclass[conference]{IEEEtran}
\IEEEoverridecommandlockouts
\usepackage{cite}
\usepackage{amsmath,amssymb,amsfonts}
\usepackage{algorithmic}
\usepackage{graphicx}
\usepackage{float}
\usepackage{textcomp}
\usepackage[usenames,dvipsnames,svgnames,table]{xcolor}
\def\BibTeX{{\rm B\kern-.05em{\sc i\kern-.025em b}\kern-.08em
    T\kern-.1667em\lower.7ex\hbox{E}\kern-.125emX}}

\usepackage{framed}
\definecolor{shadecolor}{RGB}{180,180,180}

\usepackage{array}
\usepackage{gensymb}
\usepackage[utf8]{inputenc}
\usepackage{subcaption}
\usepackage{comment}
\usepackage{balance}

\usepackage[colorlinks=true,pdfpagemode=UseNone,citecolor=black,linkcolor=black,urlcolor=BrickRed]{hyperref}
\usepackage{cleveref}

\begin{document}

% \title{Replace Linear Velocity with Angular Momentum about Contact Point in Bipedal Robot Control: Validation in A LIP-based Controller\\
% \thanks{Funding for this work was provided in part by the Toyota Research Institute (TRI) under award number No.~02281 and in part by NSF Award No.~1808051. All opinions are those of the authors.}
% }

\title{Angular Momentum about the Contact Point for Control of Bipedal Locomotion: Validation in a LIP-based Controller\\
\thanks{Funding for this work was provided in part by the Toyota Research Institute (TRI) under award number No.~02281 and in part by NSF Award No.~1808051. All opinions are those of the authors.}
}

\author{Yukai Gong and Jessy Grizzle
\thanks{The authors are with the College of Engineering and the Robotics Institute, University of Michigan, Ann Arbor, MI 48109 USA {\tt\small \{ykgong,grizzle\}}@umich.edu }
}

% \and
% \and
% \and
% \and
% \and

\maketitle

% \begin{abstract}
% Linear velocity was widely accepted as a surrogate to represent walking status in bipedal locomotion. In this paper we claim angular momentum about contact point is a better surrogate in several aspects. We implement a LIP-based controller with the new surrogate on a 20 degree-of-freedom robot, Cassie Blue, whose single leg weighs nearly one-third of the total mass. The robot achieved fast walking, turning while walking, large disturbance rejection and locomotion on rough terrain.
% \end{abstract}

\begin{abstract}
In the control of bipedal locomotion, linear velocity of the center of mass has been widely accepted as a primary variable for summarizing a robot's state vector. The ubiquitous massless-legged linear inverted pendulum (LIP) model is based on it. In this paper, we argue that angular momentum about the contact point has several properties that make it superior to linear velocity for feedback control. So as not to confuse the benefits of angular momentum with any other control design decisions, we first reformulate the standard LIP controller in terms of angular momentum. We then implement the resulting feedback controller on the 20 degree-of-freedom bipedal robot, Cassie Blue, where each leg accounts for nearly one-third of the robot's total mass of 35~Kg. Under this controller, the robot achieves fast walking, rapid turning while walking, large disturbance rejection, and locomotion on rough terrain. The reasoning developed in the paper is applicable to other control design philosophies, whether they be Hybrid Zero Dynamics or Reinforcement Learning.
\end{abstract}

% \begin{IEEEkeywords}
% \jwg{remove page numbers in final paper}
% \end{IEEEkeywords}
\pagestyle{plain}

\section{Introduction}
% Balancing is the most critical problem in bipedal locomotion. However to form a control problem it need to be quantified such that a feedback design can be possible. Many quantities are adopted as primary control variables to balance a robot, such as Center of Mass (COM) velocity\cite{kajita20013d}\cite{da20162d}\cite{Yukai2018}, COM position\cite{da2019combining}, Capture Point\cite{pratt2006capture}\cite{englsberger2011bipedal}, stance leg angle\cite{grizzle2001asymptotically}, Zero Moment Point (ZMP)\cite{1241826} and Angular Momentum\cite{GriffinIJRR2016}\cite{powell2016mechanics}.

% A bipedal robot is usually a high dimension nonlinear model. In \cite{murphy1985trotting}, a model-free control method is proposed. In \cite{KATA91}, Linear Inverted Pendulum Model is invented. SLIP model was studied in \cite{BL89}. In\cite{da2019combining},\cite{GRWE2008},\cite{reher2016realizing} full order models are utilized.

% For Bipedal locomotion, the basic balancing mechanism can be classified as foot placement\cite{kajita20013d}\cite{8968162}, touch down time control(step time control)\cite{KATA91}, ZMP control\cite{englsberger2011bipedal}, Virtual Constraint\cite{grizzle2001asymptotically}, centroidal angular momentum control\cite{1308858}\cite{xiong2020sequential}. and COM height control\cite{powell2016mechanics}.

Maintaining ``balance'' is widely viewed as the most critical problem in bipedal locomotion. The notion of ``balance'' needs to be quantified so that it can be transformed into a feedback control objective. Some represent ``balance'' with an asymptotically stable periodic orbit\cite{GRAB01,Ames2012Dynamically}. A common approach is to summarize the status of a nonlinear high-dimensional robot model with a few key variables.
The most frequently proposed variables as surrogates for ``balance'' include Center of Mass (COM) velocity\cite{kajita20013d,da20162d,HartleyGrizzleCCTA2017,Haribexo2018,Yukai2018}, COM position\cite{da2019combining}, Capture Point\cite{pratt2006capture,englsberger2011bipedal}, Zero Moment Point \cite{1241826,7363473}, and Angular Momentum\cite{GriffinIJRR2016,powell2016mechanics}.

% Some of the desirable properties of angular momentum in bipedal locomotion have been highlighted and exploited for feedback control in \jwg{include all relevant Powell papers.} \cite{grizzle2005nonlinear,westervelt2007feedback, PRTE06, powell2016mechanics, GriffinIJRR2016}. \textit{The view we take here is considerably different from these references in that we include an explicit prediction component in the feedback controller}. As in Powell and Ames \cite{??}, we chose to demonstrate our results by reformulating the well-known Linear Inverted Pendulum (LIP) model \cite{KATA91} in terms of angular momentum about the contact point. However, where Powell and Ames must adjust step frequency and the vertical velocity of the center of mass to regulate the angular momentum at the \textbf{beginning of the next step}, we are able to keep step frequency and vertical velocity of the center of mass constant, and control swing leg touch down position to achieve a \textbf{predicted value of angular momentum at the end of the next step}. 

In this paper we choose to regulate angular momentum about the contact point step-to-step as our primary control objective. Some of angular momentum's desirable properties in bipedal locomotion have been highlighted and exploited for feedback control in \cite{grizzle2005nonlinear,westervelt2007feedback, PRTE06,  GriffinIJRR2016,sano1990realization}. We choose to demonstrate our results using a reformulation of the Linear Inverted Pendulum (LIP) model \cite{KATA91} in terms of angular momentum about the contact point, as in \cite{powell2016mechanics,powell2016mechanics_2}. We emphasize that the LIP model, when reformulated in terms of angular momentum, has higher fidelity when applied to realistic robot models, than when based on linear velocity. Powell and Ames \cite{powell2016mechanics} developed a similarly reformulated LIP model and they chose to regulate the angular momentum at the beginning of the next step through touch down timing and transfer of momentum at impact. Here we take advantage of the higher fidelity of predicted angular momentum about the contact point and choose to regulate it at the end of next step, which can then be more effectively controlled by foot placement \cite{RAI84,townsend1985biped}.

To demonstrate that our results transfer in practice to a realistic bipedal robot, we implement the resulting feedback controller on the 20 degree-of-freedom bipedal robot, Cassie Blue, where each leg accounts for nearly one-third of the robot's total mass of 32~Kg. In experiments, Cassie Blue is able to execute walking in a straight line up to 2.1 m/s, simultaneously walking forward and diagonally on grass at 1 m/s, make quick, sharp turns, and handle very challenging undulating terrain. For the purpose of completeness, we note that a LIP-inspired controller organized around COM velocity has been implemented on a Cassie-series robot in \cite{xiong2019orbit}.
%We understand that it has attained walking at 1.5 m/s.

\begin{figure}[h]
    \centering
    \includegraphics[width=0.4\textwidth]{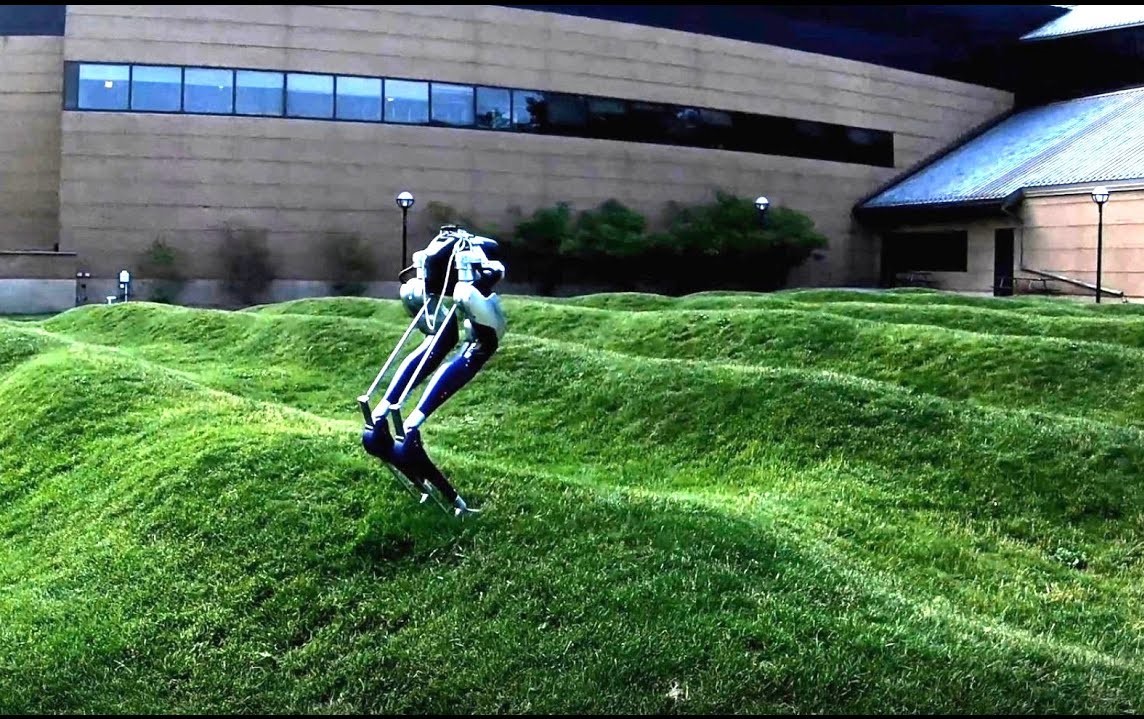}
    \caption{Cassie Blue, by Agility Robotics, on the iconic University of Michigan Wave Field.}
    \label{fig:Cassie_WaveField}
\end{figure}

The main contributions of the paper are as follows:
\begin{itemize}
    \item Demonstrate that the one-step-ahead prediction of angular momentum about the contact point provided by a LIP model is superior to a one-step-ahead prediction of linear velocity of the center of mass when applied to realistic robots;
    \item Formulate a foot placement strategy based on the one-step-ahead prediction of angular momentum.
    \item Demonstrate the resulting controller can achieve highly dynamic gaits on a 3D bipedal robot with legs that are far from massless; see Fig. \ref{fig:Cassie_WaveField}
\end{itemize}

% Heuristic model. Without the knowledge of exact model, a robot can be stabilized by heuristic tuning.\cite{RAI84}

% LIP model assumes massless leg and flat trajectory of CoM, which has linear dynamic an make it easy to analyze.\cite{KATA91}  (sLip model?)

% A Full order model captures the dynamics of a robot accurately and allows optimization. you
% Often the dynamic is analyzed on a low dimension manifold. \cite{WESTERVELTE03} defines virtual constraints for joints of a robot with a mechanical phase. \cite{da2019combining} Propose virtual constraints for strongly actuated joints with weakly actuated joints and time. \cite{xie2018feedback,castillo2019reinforcement} use reinforcement learning method to generate controller for biped robot.

% \cite{westervelt2007feedback,PRTE06} Discussed some properties of angular momentum 

% \cite{GriffinIJRR2016} create non-holonomic constraint w.r.t. to angular momentum about contact point. 

\begin{figure} 
    \centering
    \includegraphics[width=0.5\textwidth]{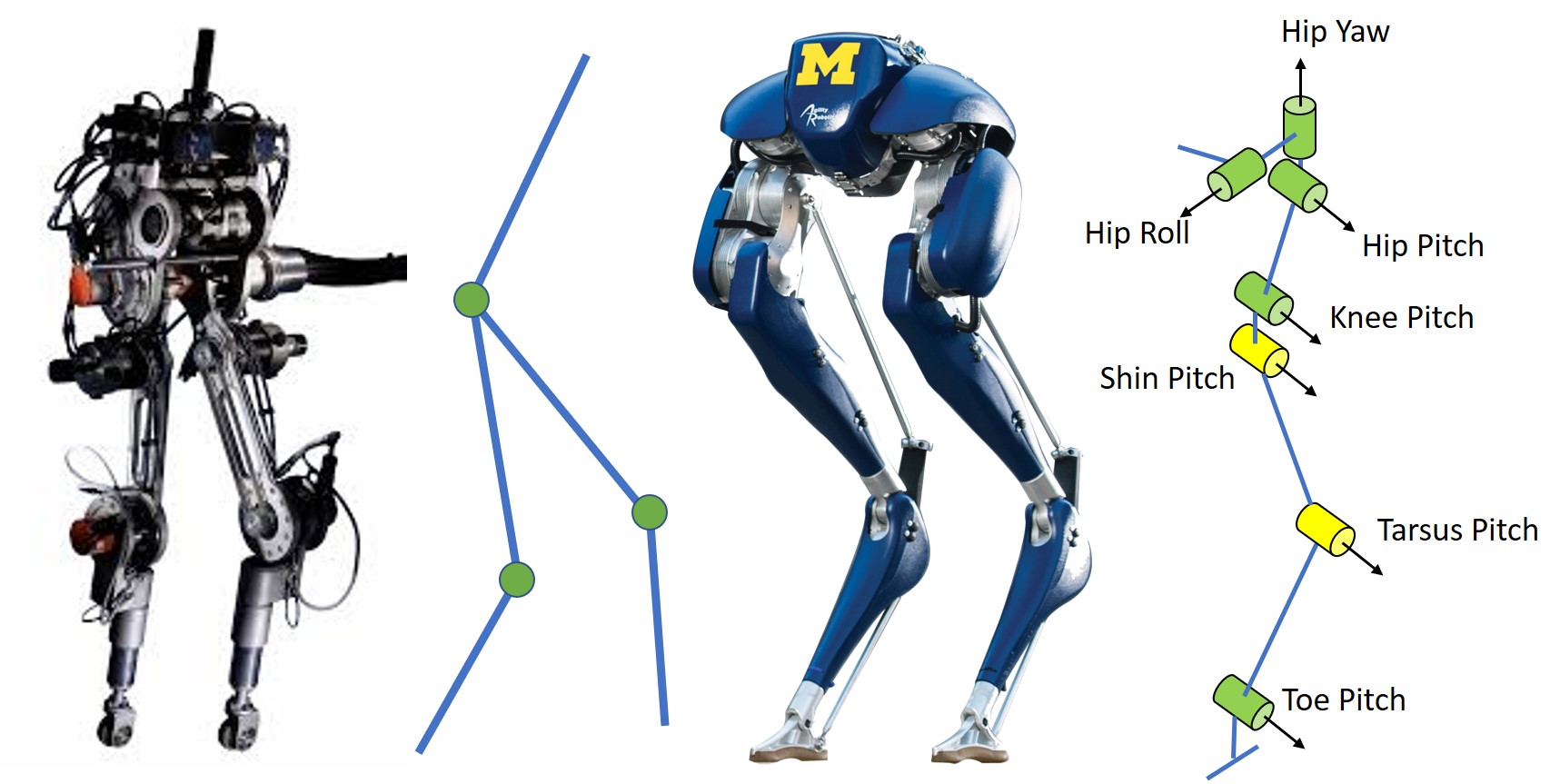}
    \caption{Rabbit and Cassie. Rabbit is planar robot with 2 joints on each leg and Cassie is 3D robot with 7 joints on each leg. Actuated joints are highlighted in green, passive in yellow.}
    \label{fig:Rabbit_Cassie}
\end{figure}

We will use both Rabbit \cite{CHABAOPLWECAGR02} and Cassie Blue to illustrate our developments in the paper. Experiments will be conducted exclusively on Cassie. 
Rabbit is a 2D biped with five links, four actuated joints, and a mass of 32~Kg; see Fig.~\ref{fig:Rabbit_Cassie}. Each leg weighs 10 kg, with 6.8 kg on each thigh and 3.2 kg on each shin. 

The bipedal robot shown in Fig.~\ref{fig:Cassie_WaveField}, named Cassie Blue, is designed and built by Agility Robotics. The robot weighs 32kg. It has 7 deg of freedom on each leg, 5 of which are actuated by motors and 2 are constrained by springs; see Fig. \ref{fig:Rabbit_Cassie}. A floating base model of Cassie has 20 degrees of freedom. Each foot of the robot is blade-shaped and provides 5 holonomic constraints when it is on ground. Though each of Cassie's legs has approximately 10 kg of mass, most of the mass is concentrated on the upper part of the leg. In this regard, the mass distribution of Rabbit is a bit more typical of bipedal robots, which is why we include the Rabbit model in the paper.

% \begin{figure} 
%     \centering
%     \includegraphics[width=0.45\textwidth]{Graphics/Rabbit.png}
%     \caption{Rabbit is an underactuated planar bipedal robot with point feet. Its floating base model has seven degrees of freedom and four actuators.}
%     \label{fig:Rabbit_Pic}
% \end{figure}

% \begin{figure} 
%     \centering
%     \includegraphics[width=0.4\textwidth]{Graphics/Kinematics_2.jpg}
%     \caption{ Cassie bipedal robot along with the kinematics of one leg. The five green joints are actuated and two yellow joints are constrained by very rigid springs. When springs deflections are 0. Shin Pitch angle is 0 and Tarsus Pitch angle is negative Knee Pitch angle plus a 13 degree offset. Notice: In following text, toe refer to the position of toe joint.}
%     \label{fig: Cassie_Kinematics}
% \end{figure}

The remainder of the paper is organized as follows. Section \ref{Sec:Angular Momentum} and \ref{Sec:LIP} introduce angular momentum and the LIP model. In Section \ref{Sec:Controller}, we show how to predict the evolution of angular momentum with a LIP model and how to use the prediction to decide foot placement. This provides a  feedback controller that will stabilize a 3D LIP. In Section \ref{sec:ImplementOnCassie}, we provide our path to implementing the  controller on  Cassie Blue. Additional reference trajectories are required beyond a path for the swing foot, and we provide ``an intuitive'' method for their design. Section \ref{Sec:Experiment} addresses several practical challenges associated with the passage from simulation to experimental implementation on Cassie. Section \ref{Sec:ExperimentResults} shows the experiment results. Conclusion are give in Sect.~\ref{Sec:Conclusion}.

\section{Angular Momentum About Contact Point} \label{Sec:Angular Momentum}
% \begin{snugshade*}
% This section discusses the good properties of angular momentum about contact point ($L$)
% \end{snugshade*}
Some of the properties of angular momentum in bipedal locomotion have been discussed in\cite{westervelt2007feedback, PRTE06, grizzle2005nonlinear}, and in \cite{GriffinIJRR2016}, non-holonomic virtual constraints are created with angular momentum. The view we take here is considerably different for these references. In the following, we will address two questions: why we can replace linear momentum with angular momentum for the design of feedback controllers and what are the benefits of doing so.  %Adjustments for feet will be discussed later. 

Initially, we address the \textit{single support} phase of walking, meaning only one leg is in contact with the ground. Moreover, we are considering a point contact.

Let $L$ denote the \textbf{angular momentum} about the contact point of the stance leg. The relationship between angular momentum and \textbf{linear momentum} for a 3D bipedal robot is
\begin{equation}
\label{eq:AngularMomentumLinearMomentum}
    L = L_{\rm CoM} + p \times m_{\rm tot} v_{\rm CoM},
\end{equation}
where $L_{\rm CoM}$ is the angular momentum about the center of mass, $v_{\rm CoM}$ is the linear velocity of the center of mass, $m_{\rm tot}$ is the total mass of the robot, and $p$ is the vector emanating from the contact point to the center of mass.

 For a bipedal robot that is walking instead of doing somersaults, the angular momentum about the center of mass must oscillate about zero. Hence, \eqref{eq:AngularMomentumLinearMomentum} implies that the difference between $L$ and $p\times m v_{\rm CoM}$ also oscillates around zero, which we will write as
 \begin{equation}
\label{eq:AngularMomentumLinearMomentum02}
  L - p \times m_{\rm tot} v_{\rm CoM}  = L _{\rm CoM} \text{~oscillates about~} 0.
\end{equation}
From \eqref{eq:AngularMomentumLinearMomentum02}, we see that we approximately obtain a desired linear velocity by regulating $L$. 

The discussion so far has focused on a single support phase of a walking gait. Bipedal walking is characterized by the transition between left and right legs as they alternately take on the role of stance leg (aka support leg) and swing leg (aka non-stance leg). In double support, the transfer of angular momentum between the two contact points satisfies
% \begin{equation}\label{AM_transfer}
%     L_{\rm sw} =  L_{\rm st} + p_{\rm sw2st}\times m_{\rm tot} v_{\rm CoM}
% \end{equation}
\begin{equation}\label{AM_transfer}
    L_{2} =  L_{1} + p_{2\to 1}\times m_{\rm tot} v_{\rm CoM}
\end{equation}
where $L_i$ is the angular momentum about contact point $i$ and $p_{1 \to 2}$ is the vector from contact point $1$ to contact point $2$. 

Hence, one can replace the control of linear velocity with control of angular momentum about the contact point. \textbf{But what are the advantages?}
\begin{enumerate}
\renewcommand{\labelenumi}{(\alph{enumi})}
\setlength{\itemsep}{.2cm}
    \item The first advantage of controlling $L$ is that it provides a more comprehensive representation of current walking status by including both $L_{ \rm CoM}$ and $p\times mv_{\rm CoM}$, between which momentum moves forth and back during a step.
    \item Secondly, $L$ has a relative degree three with respect to motor torques, if ankle torque is zero. Indeed, in this case,
\begin{equation} 
\label{eq:AngularMomentumDot}
    \dot{L} = p \times m_{\rm tot}g.
\end{equation}
where $g$ is the gravitational constant. Consequently, $L$ is very weakly affected by peaks in motor torque that often occur in off nominal conditions. Moreover, if a limb, such as the swing leg, is moving quickly in response to a disturbance, it will strongly affect the angular momentum about the center of mass and the robot's linear velocity, while leaving the angular momentum about the contact point only weakly affected.
\item Thirdly, $\dot{L}$ is ONLY a function of the center of mass position, making it easy to predict its trajectory over a step. 
\item Angular momentum about a given contact point is invariant under impacts at that point, and the change of angular momentum between two contact points depends only on the vector defined by the two contact points and the CoM velocity. Hence, we can easily determine the angular momentum about the new contact point by \eqref{AM_transfer} when impact happens without approximating assumptions about the impact model. Moreover, if the vertical component of the $v_{\rm CoM}$ is zero and the ground is level, then $p_{2\to 1}\times m_{\rm tot} v_{\rm CoM}=0$  and hence $L_2 = L_1$.
\end{enumerate}

\begin{figure*}
\centering
\begin{subfigure}{.32\textwidth}
  \centering
  \includegraphics[width=\textwidth]{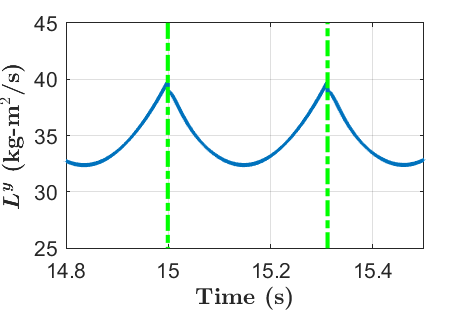}
  \caption{Rabbit $L^y$}
\end{subfigure}
\begin{subfigure}{.32\textwidth}
  \centering
  \includegraphics[width=\textwidth]{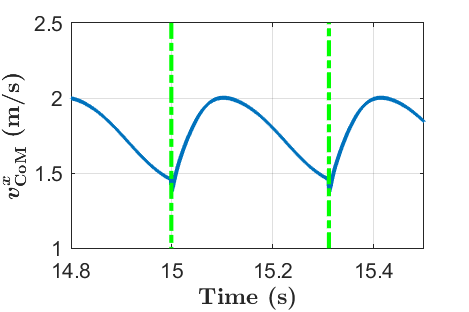}
  \caption{Rabbit $v^x_{\rm CoM}$}
\end{subfigure}
\begin{subfigure}{.32\textwidth}
  \centering
  \includegraphics[width=\textwidth]{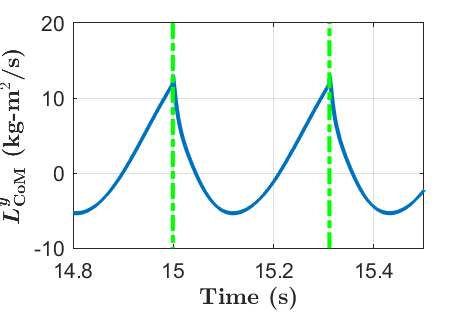}
  \caption{Rabbit $L^y_{\rm CoM}$}
\end{subfigure}

\begin{subfigure}{.32\textwidth}
  \centering
  \includegraphics[width=\textwidth]{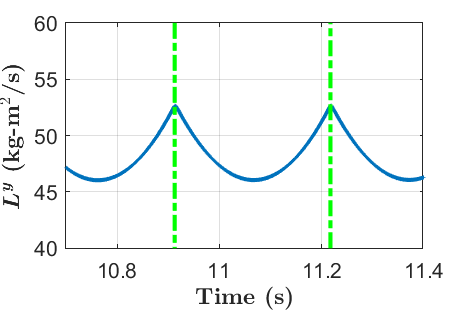}
  \caption{Cassie $L^y$}
\end{subfigure}
\begin{subfigure}{.32\textwidth}
  \centering
  \includegraphics[width=\textwidth]{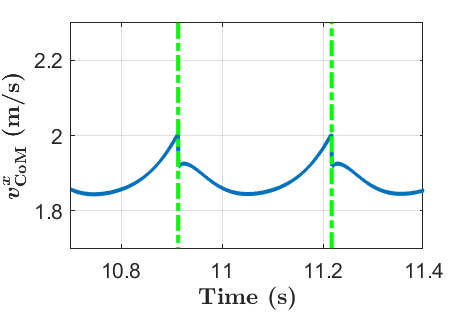}
  \caption{Cassie $v^x_{\rm CoM}$}
\end{subfigure}
\begin{subfigure}{.32\textwidth}
  \centering
  \includegraphics[width=\textwidth]{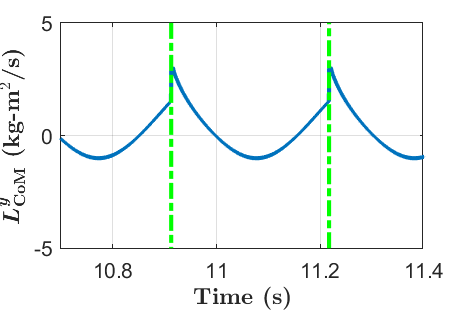}
  \caption{Cassie $L^y_{\rm CoM}$}
\end{subfigure}

\caption{Comparison of $L$, $L_{\rm CoM}$, $v^x_{\rm CoM}$ in simulation for the bipedal robots Rabbit and Cassie, while $v^z_{\rm CoM}$ is carefully regulated to zero. The angular momentum about the contact point, $L$, has a convex trajectory, both of which are similar to the trajectory of a LIP model,  while the trajectory of the longitudinal velocity of the center of mass, $v^x_{\rm CoM}$, has no particular shape. In this figure, $L$ is continuous at impact, which is based on two conditions: $v^z_{\rm CoM}=0$ at impact and the ground is level. Even when these two conditions are not met, the jump in $L$ at impact can be easily calculated with \eqref{AM_transfer}.}
\label{fig:comparison_of_L_V}
\end{figure*}

Figure \ref{fig:comparison_of_L_V} shows simulation plots of $L$, $L_{\rm CoM}$, and $v^x_{\rm CoM}$ for the planar bipedal robot, Rabbit, and the 3D bipedal robot, Cassie Blue. It is seen that the angular momentum about the contact point has the advantages discussed above.

\section{Linear Inverted Pendulum Model}\label{Sec:LIP}

This section provides the ubiquitous Linear Inverted Pendulum (LIP) model of 
Kajita et al.~\cite{KATA91}. The LIP model assumes the center of mass moves in a plane, the angular momentum about the center of mass is constant, and the legs are massless. Here, we will express the model in terms of its original  coordinates, namely position and linear velocity of the center of mass, and in terms of the proposed new coordinates, namely position and angular momentum about the contact point. The dynamics of the inverted pendulum are exactly linear. Moreover, the 3D dynamics in the $x$ and $y$ directions are decoupled, and hence we only need to consider a $2D$ pendulum.

\begin{figure}
    \centering
    \includegraphics[width=0.25\textwidth]{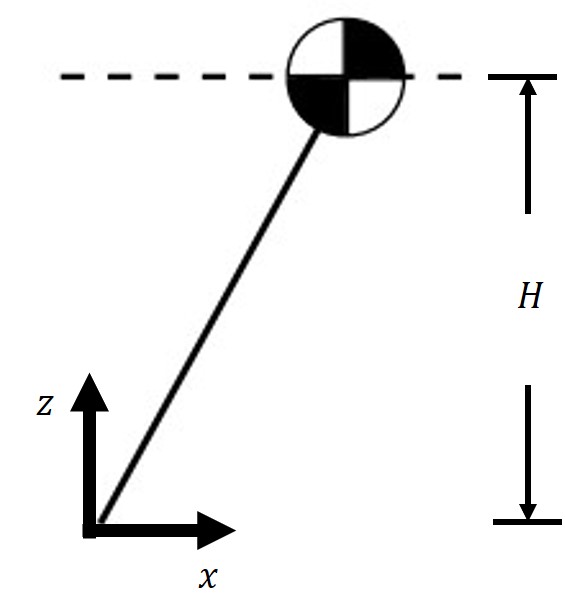}
    \caption{Linear Inverted Pendulum Model. A prismatic joint in the leg allows the CoM to move along a given line. In the original paper \cite{KATA91}, the model requires massless legs, constant $L_{COM}$, and the center of mass moving along a straight (not necessarily horizontal) line. Here, we assume in addition a point mass and a horizontal trajectory for the center of mass.}
    \label{fig:my_label}
\end{figure}

Let $H$ denote the height of the center of mass. For a 2D model, the dynamics in the $x$ direction is
\begin{equation} \label{eq:LIP_original_eqn}
\begin{bmatrix}
\dot{x} \\
\ddot{x} \\
\end{bmatrix}
=
\begin{bmatrix}
0 & 1 \\
g/H & 0\\
\end{bmatrix}
\begin{bmatrix}
x \\
\dot{x}
\end{bmatrix},
\end{equation}
where x is the position of CoM in the frame of contact point.
if we assume there is no ankle torque. The solution of this linear system is
% \begin{equation} \label{eq:LIP_original_sol}
% \begin{bmatrix}
% x(t) \\
% \dot{x}(t) \\
% \end{bmatrix}
% =
% \begin{bmatrix}
% \cosh(\ell t) & 1/\ell\sinh(\ell t) \\
% \ell\sinh(\ell t) & \cosh(\ell t)
% \end{bmatrix}
% \begin{bmatrix}
% x_0 \\
% \dot{x}_0
% \end{bmatrix},
% \end{equation}
% where $\ell = \sqrt{\frac{g}{H}}$. From this equation, as estimate of the state variables at the end of a step in terms of their current values is given by
\begin{equation} \label{eq:LIP_original_Preiction}
\begin{bmatrix}
x(T) \\
\dot{x}(T) \\
\end{bmatrix}
=
\begin{bmatrix}
\cosh(\ell (T-t)) & 1/\ell \sinh(\ell (T-t)) \\
\ell \sinh(\ell (T-t)) & \cosh(\ell (T-t))
\end{bmatrix}
\begin{bmatrix}
x(t) \\
\dot{x}(t)
\end{bmatrix},
\end{equation}
where $\ell = \sqrt{\frac{g}{H}}$, $t$ is the current time and $T$ is the (predicted) time of the end of the step.

We assume the body is a point mass and is moving on a horizontal plane, that is the height of the center of mass is constant. Because we are assuming a point mass, the angular momentum about the center of mass is zero. We now replace the states $\{x,\dot{x}\}$ with $\{x,L^y\}$, where $L^y$ is the y-component of angular momentum about the contact point. The corresponding dynamic model is 
\begin{equation} \label{eq:LIP_AM_eqn}
\begin{bmatrix}
\dot{x} \\
\dot{L}^y \\
\end{bmatrix}
=
\begin{bmatrix}
0 & 1/mH \\
mg & 0\\
\end{bmatrix}
\begin{bmatrix}
x \\
L^y
\end{bmatrix},
\end{equation}
and its corresponding solution is
% \begin{equation} \label{eq:LIP_AM_sol}
% \begin{bmatrix}
% x(t) \\
% L^y(t) \\
% \end{bmatrix}
% =
% \begin{bmatrix}
% \cosh(\ell t) & 1/mH\ell\sinh(\ell t) \\
% mH\ell\sinh(\ell t) & \cosh(\ell t)
% \end{bmatrix}
% \begin{bmatrix}
% x_0 \\
% L_0^y
% \end{bmatrix}.
% \end{equation}
% We can also rewrite \eqref{eq:LIP_AM_sol} in the form of a predictor,
\begin{equation} \label{eq:LIP_AM_Preiction}
\small
\begin{bmatrix}
x(T) \\
L^y(T) \\
\end{bmatrix}
=
\begin{bmatrix}
\cosh(\ell(T-t)) & 1/mH\ell\sinh(\ell(T-t)) \\
mH\ell\sinh(\ell(T-t)) & \cosh(\ell(T-t))
\end{bmatrix}
\begin{bmatrix}
x(t) \\
L^y(t)
\end{bmatrix},
\end{equation}
where $t$ is the current time and $T$ is the (predicted) time of the end of the step. 

\begin{figure*}
\centering
\begin{subfigure}{.48\textwidth}
  \centering
  \includegraphics[width=\textwidth]{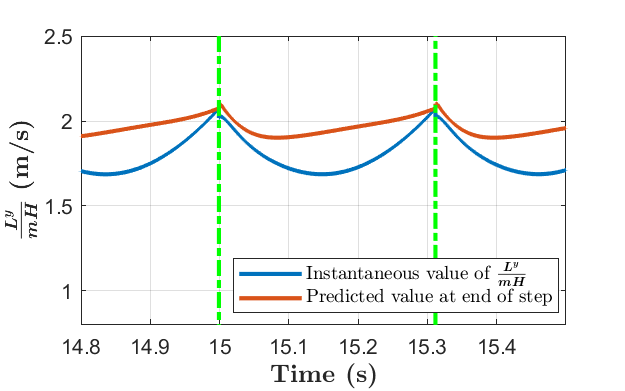}
  \caption{Rabbit $L^y$ prediction}
  \label{fig:L_prediction_Rabbit}
\end{subfigure}
\begin{subfigure}{.48\textwidth}
  \centering
  \includegraphics[width=\textwidth]{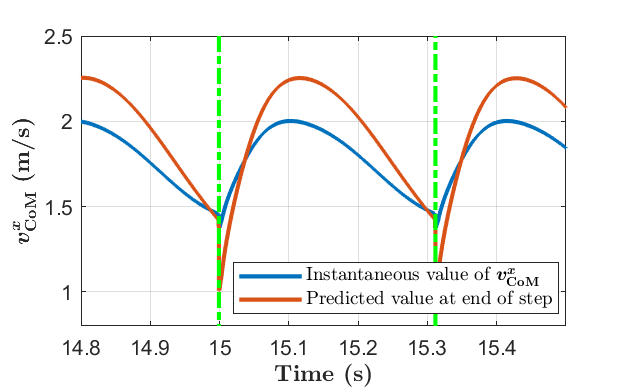}
  \caption{Rabbit $v^x_{\rm CoM}$ prediction}
  \label{fig:vx_prediction_Rabbit}
\end{subfigure}

\begin{subfigure}{.48\textwidth}
  \centering
  \includegraphics[width=\textwidth]{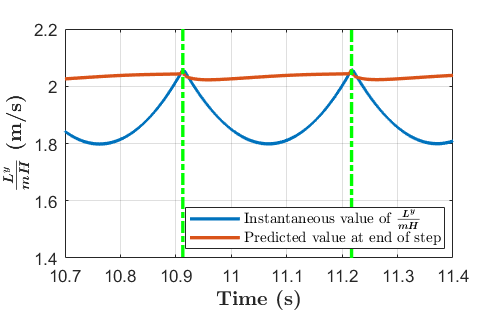}
  \caption{Cassie $L^y$ prediction}
  \label{fig:L_prediction_Cassie}
\end{subfigure}
\begin{subfigure}{.48\textwidth}
  \centering
  \includegraphics[width=\textwidth]{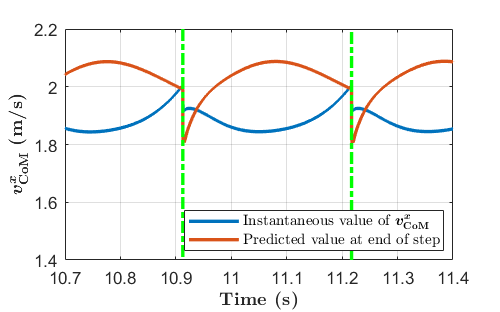}
  \caption{Cassie $v^x_{\rm CoM}$ prediction}
  \label{fig:vx_prediction_Cassie}
\end{subfigure}

\caption{Comparison of the ability to predict velocity vs angular momentum at the end of a step. The most crucial decision in the control of a bipedal robot is where to place the next foot fall. In the standard LIP controller, the decision is based on predicting the longitudinal velocity of the center of mass. In Sect.~\ref{Sec:LIP} we use angular momentum about the contact point. We do this because on realistic bipeds, the LIP model provides a more accurate and reliable prediction of $L$ than $v_{\rm CoM}$. The comparison is more significant on Rabbit, whose leg center of mass is further away from the overall CoM.}
\label{fig:LIP_prediction}
\end{figure*}

For a point-mass inverted pendulum, where the mass moves on a horizontal plane, representations \eqref{eq:LIP_original_eqn} and \eqref{eq:LIP_AM_eqn} are exactly the same. \textbf{So what have we gained?} \textbf{Importantly, for a real robot, where the two representations are only approximate, the second one is better for making predictions on the robot's state}, as we discussed in Sec.~\ref{Sec:Angular Momentum}. 
Figure~\ref{fig:LIP_prediction} compares the predictions of linear velocity and angular velocity about the contact point for a seven degree of freedom 2D model of Rabbit and a 20 degree of freedom 3D model of Cassie. In the figure, the simulated instantaneous values of $v^x_{\rm CoM}(t)$ and $L^y(t)$ are shown in blue. The red line shows the evolution of the predicted values at the end of a step for $v^x_{\rm CoM}(t)$ and $L^y(t)$ from \eqref{eq:LIP_original_Preiction} and \eqref{eq:LIP_AM_Preiction}  plotted against the corresponding instantaneous values. In a perfect predictor, the predicted values would be straight lines. It is clear that, when extrapolated to a realistic model of a robot, the prediction of angular momentum about the contact point is significantly more reliable than the estimate of linear velocity.

\section{High-Level Control Strategy in Terms of Angular Momentum} \label{Sec:Controller}
% \begin{snugshade*}
% This section discusses how to use LIP model to predict the evolution of $L$, and decides the foot placement based on the prediction of $L$. And the generation of reference trajectory.
% \end{snugshade*}
Our overall control objective will be to regulate walking speed.
Because of the four advantages of angular momentum versus linear velocity that we listed in Sect.~\ref{Sec:LIP}, we will use angular momentum about the contact point as a primary control variable. In this section, we explain our method for deciding where to end one step by initiating contact between the ground and the swing foot, thereby beginning the next step. In robot locomotion control, this is typically called ``foot placement control''. 

\subsection{Notation}

We need to distinguish among the following time instances when specifying the control variables. 
\begin{itemize}
    \item $T$ is the step time.
    \item $T_k$ is the time of the $k$th impact. 
    \item $T_k^-$ is the end time of step $k$, so that
    \item $T_k^+$ is the beginning time of step $k+1$ and $T_{k+1}^-$ is the end time of step $k+1$.
    \item $(T_k^- -t)$ is the time until the end of step $k$.
\end{itemize}
The superscripts $+$ and $-$ on $T_k$ are due to the impact map; see \cite{westervelt2007feedback}.

\begin{itemize}
    \item $p_{\rm st \to CoM}$. Vector emanating from stance foot to CoM. Here, the stance foot can be thought of as the current contact point.
    \item $p_{\rm sw \to CoM}$. Vector emanating from swing foot to CoM.  Here, the swing foot is defining the of contact for the next impact and hence will be a control variable.
\end{itemize}

\begin{figure}
    \centering
    \includegraphics[width=0.5\textwidth]{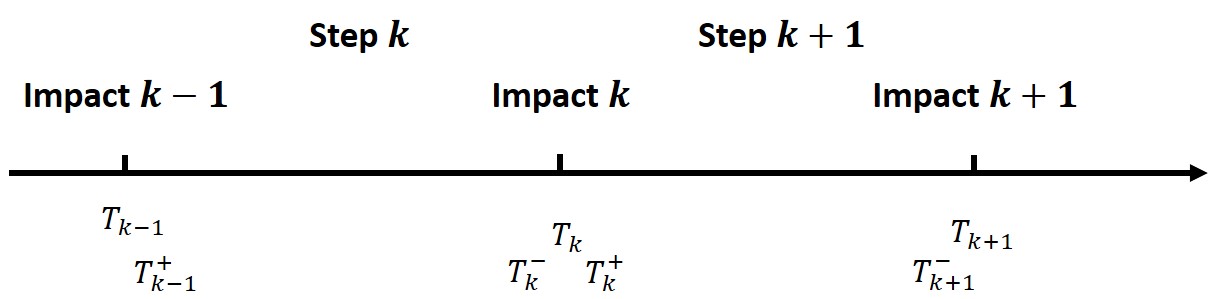}
    \caption{Definition of $T_k$}
    \label{fig:Def_Time}
\end{figure}

\begin{figure}
    \centering
    \includegraphics[width=0.4\textwidth]{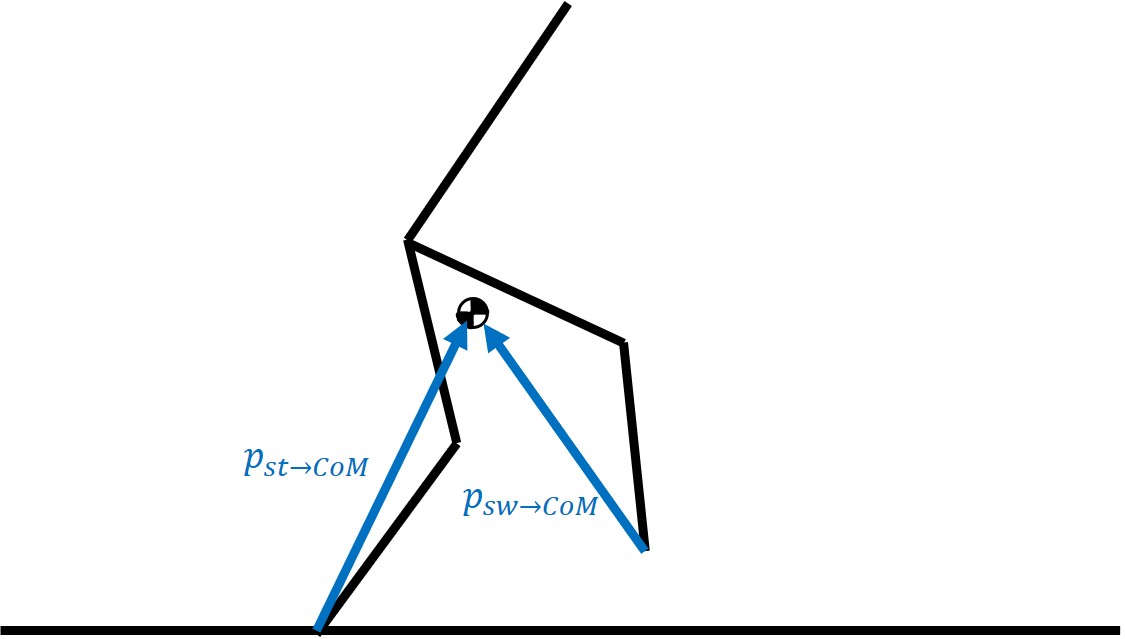}
    \caption{Definition of $p_{\rm st \to CoM}$ and $p_{\rm sw \to CoM}$}
    \label{fig:Def_p}
\end{figure}
\subsection{Foot placement in longitudinal direction}
There are multiple means to stabilize gaits\cite{westervelt2007feedback}. Because we seek to focus on the benefits of using angular momentum instead of linear velocity, the controller we give here simply uses ``foot placement'' as is commonly associated with the LIP model. \textbf{The principal idea is to regulate the angular momentum at the end of the \textit{next} step by choosing the foot placement position for the end of the \textit{current} step.}  

The Angular Momentum at the end of the next step is related to the Angular Momentum and the position of the center of mass at the beginning of next step by
\begin{equation} \label{eq:AM_next_end_predict}
{L}^y(T_{k+1}^-) = mH\ell\sinh(\ell T)p^x_{\rm st \to CoM}(T_k^+) + \cosh(\ell T){L}^y(T_k^+).
\end{equation}
However, the CoM position with respect to the stance foot at the beginning of the next step is the same as the CoM position with respect to the swing foot at the end of the current step,
\begin{equation} \label{eq:rp_impact_1}
    p^x_{\rm st \to CoM}(T_k^+) = p^x_{\rm sw \to CoM}(T_k^-).
\end{equation}
Because $p^x_{\rm sw \to CoM}(T_k^-)$ is what we can control in the current step, 
we are motivated to re-write \eqref{eq:AM_next_end_predict} as
\begin{equation} \label{eq:AM_next_end_predictv02} \small
{L}^y(T_{k+1}^-) = mH\ell*\sinh(\ell T)p^x_{\rm sw \to CoM}(T_k^-) + \cosh(\ell T){L}^y(T_k^+).
\end{equation}

Solving \eqref{eq:AM_next_end_predictv02} for the desired foot position would not yield a causal formula. However, because the CoM height is assumed constant and the ground is flat, the angular momentum about the next contact point is equal to the angular momentum about the current stance leg,
\begin{equation} \label{eq:AM_impact_relation_1}
   {L}^y(T_k^+) = {L}^y(T_k^-);
\end{equation}
see \eqref{AM_transfer}. This yields
\begin{equation} \label{eq:AM_next_end_predictv03} \small
{L}^y(T_{k+1}^-) = mH\ell\sinh(\ell T)p^x_{\rm sw \to CoM}(T_k^-) + \cosh(\ell T){L}^y(T_k^-).
\end{equation}

Now, we are almost there. From the second row of \eqref{eq:LIP_AM_Preiction}, an estimate for the angular momentum about the contact point at the end of current step, $\widehat{L}^y(T_k^-)$, can be continuously estimated by
\begin{equation} \label{eq:AM_this_end_predict} \small
    \widehat{L}^y(T_k^-)(t) = m H \ell \sinh(\ell(T_k^- - t))p^x_{\rm st}(t) + \cosh(\ell (T_k^- - t))L^y(t).
\end{equation}

If ${L}^y(T_{k+1}^-)$ is replaced by its desired value, ${L}^{y~\rm des}(T_{k+1}^-)$, we now have a causal and implementable expression for desired foot placement,  
\begin{snugshade*}
\begin{equation} \label{eq:Desired_Footplacement}
    p^{x~{\rm des}}_{\rm \bf sw \to CoM}(T_k^-)(t) :=\frac{L^{y ~{\rm des}}(T^-_{k+1}) - \cosh(\ell T)\widehat{L}^y(T_k^-)(t)}{mH \ell \sinh(\ell T)}.
\end{equation}
\end{snugshade*}

With a perfect estimator for the angular momentum about that stance leg, such as is possible with an ideal LIP model, the desired foot position would be constant. Here, it varies because in a real robot, our estimate for angular momentum evolves with time.

\subsection{Lateral Control}
From \eqref{eq:AngularMomentumDot}, the time evolution of the angular momentum about the contact point is decoupled about the $x$- and $y$-axes. Therefore, once a desired angular momentum at the end of next step is given, Lateral Control is essentially identical to Longitudinal Control and \eqref{eq:Desired_Footplacement} can be applied equally well in the lateral direction. \textbf{The question becomes how to decide on  $\mathbf{L^{x~ \rm \bf des}(T_{k+1}^-)}$.}

For walking in place or walking with zero lateral velocity, it is sufficient to obtain $L^{x~ \rm des}$ from a periodically oscillating LIP model,
\begin{equation} \label{eq:lateral_desired_Lx}
    L^{x~ \rm des}(T_{k+1}^-) = \pm\frac{1}{2}mHW\frac{\ell\sinh(\ell T)}{1+\cosh(\ell T)},
\end{equation}
where $W$ is the desired step width. The sign is positive if next stance is left stance and negative if next stance is right stance.

% When turning, how to select values for $L^{x~{\rm des}}$ and $L^{y~{\rm des}}$ is a bit more complicated and may be robot dependent. We explain our method for Cassie Blue in the next Section.
\subsection{Turning}
In each step, we suppose there is an angle $D_k$ which defines the \textbf{forward direction} of that step. When walking in a straight line, $D_k$ is a constant. When the robot makes turns, $D_k$ will change step by step, $\Delta D_k = D_{k+1} - D_k \neq 0$. The $L^{x~{\rm des}}$ and $L^{y~{\rm des}}$  will then be first decided in their corresponding frame defined by $D_k$, then transformed back in the world frame and used to decide foot placement in the world frame.
% \subsection{Turning} \yg{top view figure,or defining Dk}

% We assume that each step of the robot can be associated to a scalar variable $D_k$, for example as defined in Fig.~\ref{??}, which specifies the direction of the robot in the ground plane. This direction also defines a frame \{$D_k$\}, whose $x$ axis is aligned with $D_k$ and $z$ axis is vertical. When robot is making a turn the direction $D_k$ and $D_{k+1}$ will be different. And the desired Angular Momentum $L(T_{k+1}^-)$  will first be decided in frame \{$D_{k+1}$\}, $L_y(T_{k}^+)$ will be predicted in frame \{$D_k$\} and then converted to frame \{$D_{k+1}$\} and finally the foot placement will be decided in frame \{$D_{k+1}$\}.

% \yg{When making prediction of $L$, it is easier to predict in the frame of stance toe because the toe torque ( or damping) will be aligned with $y$ axis. If we assume no toe torque at all, there is no need for those frames mentioned above}

\section{Implementing the LIP-based Angular Momentum Controller on a Real Robot} \label{sec:ImplementOnCassie}

In this section we introduce the control variables for Cassie Blue and generate their reference trajectory. As in \cite{Yukai2018}, we leave the stance toe passive. Consequently, there are nine (9) control variables, listed below from the top of the robot to the end of the swing leg,
\begin{equation} \label{eq:control variables}
h_0=
\begin{bmatrix}
\rm torso \; pitch \\
\rm torso \; roll \\
\rm stance \; hip \; yaw \\
\rm swing \; hip \; yaw \\
p^z_{\rm st \to CoM} \\
p^x_{\rm sw \to CoM} \\
p^y_{\rm sw \to CoM} \\
p^z_{\rm sw \to CoM} \\
\rm swing \;toe \;absolute \;pitch \\
\end{bmatrix}.
\end{equation}
For later use, we denote the value of $h_0$ at the beginning of the current step by $h_0(T^+_{k-1})$. When referring to individual components, we'll use $h_{03}(T^+_{k-1})$, for example.

We first discuss variables that are constant. The reference values for torso pitch, torso roll, and swing toe absolute pitch are constant and zero, while the reference for $p^z_{\rm st \to CoM}$, which sets the height of the CoM with respect to the ground, is constant and equal to $H$.

We next introduce a phase variable
\begin{equation} \label{eq:phase_variable}
    s := \frac{t - T_{k-1}^+}{T}
\end{equation}
that will be used to define quantities that vary throughout the step to create ``leg pumping'' and ``leg swinging''. The reference trajectories of $p^x_{\rm sw \to CoM}$ and $p^y_{\rm sw \to CoM}$ are defined such that:
\begin{itemize}
    \item at the beginning of a step, their reference value is their actual position;
    \item the reference value at the end of the step implements the foot placement strategy in \eqref{eq:Desired_Footplacement}; and
    \item in between a half-period cosine curve is used to connect them, which is similar to the trajectory of an ordinary (non-inverted) pendulum.
\end{itemize}
The reference trajectory of $p^z_{\rm sw \to CoM}$ assumes the ground is flat and the control is perfect: 
\begin{itemize}
    \item at mid stance, the height of the foot above the ground is given by $z_{CL}$, for the desired vertical clearance.
\end{itemize}

The reference trajectories for the stance hip and swing hip yaw angles are simple straight lines connecting their initial actual position and their desired final positions. For walking in a straight line, the desired final position is zero. To include turning, the final value has to be adjusted.  Suppose that a turn angle of $\Delta D^{\rm des}_k$ radians is desired. One half of this value is given to each yaw joint:
\begin{itemize}
\item $+ \frac{1}{2}\Delta D^{\rm des}_k \to \rm swing \; hip \; yaw$; and
\item $- \frac{1}{2}\Delta D^{\rm des}_k \to \rm stance \; hip \; yaw$
\end{itemize}
The signs may vary with the convention used on other robots.

\begin{figure}
    \centering
    \includegraphics[width=0.5\textwidth]{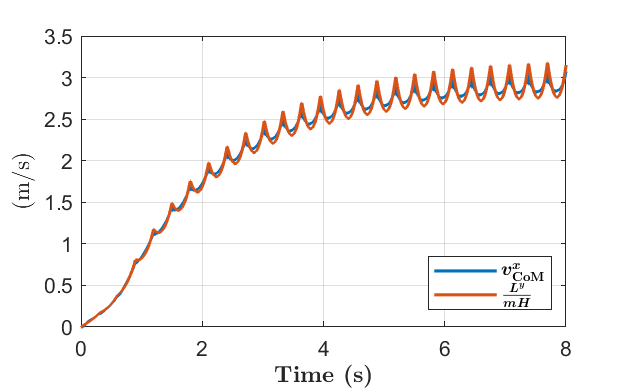}
    \caption{Simulation results of Cassie. $\frac{L^{x \; \rm des}}{mH}$ ramped up from $0$ to $3$ m/s. }
    \label{fig:sim_result}
\end{figure}

The final result for Cassie Blue is
\begin{equation} \label{eq:control variables reference} \small
\begin{aligned}
&h_{\rm d}(s):=\\
&
\begin{bmatrix}
0 \\
0 \\
(1-s)h_{03}(T_{k-1}^+) + s(-\frac{1}{2}(\Delta D_{k})) \\
(1-s)h_{04}(T_{k-1}^+) + s(\frac{1}{2}(\Delta D_{k})) \\
H \\
\frac{1}{2}\big[(1+\cos(\pi s))h_{06}(T_{k-1}^+) + (1 - \cos(\pi s))p^{x ~ \rm des}_{\rm sw \to CoM}(T_k^-)\big] \\
\frac{1}{2}\big[(1+\cos(\pi s))h_{07}(T_{k-1}^+) + (1 - \cos(\pi s))p^{y ~ \rm des}_{\rm sw \to CoM}(T_k^-)\big] \\
4 z_{cl} (s-0.5)^2+(H-z_{CL}); \\
0 \\
\end{bmatrix}
\end{aligned}.
\end{equation}
When implemented with an Input-Output Linearizing Controller so that $h_0$ tracks $h_d$, the above control policy allows Cassie to move in 3D in simulation. Fig.~\ref{fig:sim_result} shows Cassie starts from a walking in place gait and accelerate to a speed of $2.8$~m/s.
%\subsection{Turning}
%For each step, there is a target direction(denoted by $tgd$ later), ideally the stance toe should be pointing at the target direction. When turning, there will be a difference between the target direciton of \textbf{next} step and stance toe direction of \textbf{this} step. The desired angle of both hip yaw motor at the end of this step should be half of the difference. For example, if the difference is $ 30 \degree $, than the desired angle of both yaw motor should be $ \pm 15 \degree $degree, such that the swing toe is pointing in the target direction of \textbf{next} step at the end of \textbf{this} step. The reference for both yaw motors are straight lines connecting their actual position at the beginning of a step and desired position at the end of step.

%Each target direction defines a frame, where the x axis is pointing in the target direction and xy plane is horizontal. $\tilde{L}^{\rm next \;f}_{x }$ and $\tilde{L}^{\rm next\;f}_{y }$ are defined in \textbf{next} frame first and then convert back to \textbf{this} frame, which will finally decide $\widetilde{rp}^{\rm this\;f}_{\rm x\; swToe}$ in \textbf{this} frame.

% \subsection{Simulation Result} \label{Sec:Simulation}
% Cassie is able to walk at 4m/s in an ideal simulator. In the ideal simulator the springs are rigid and sensors are perfect. Control Variables track their references by an Input-Output Controller.

\section{Modifications for Practical Implementation} \label{Sec:Experiment}
\begin{figure*}
    \centering
    \includegraphics[width=0.8\textwidth]{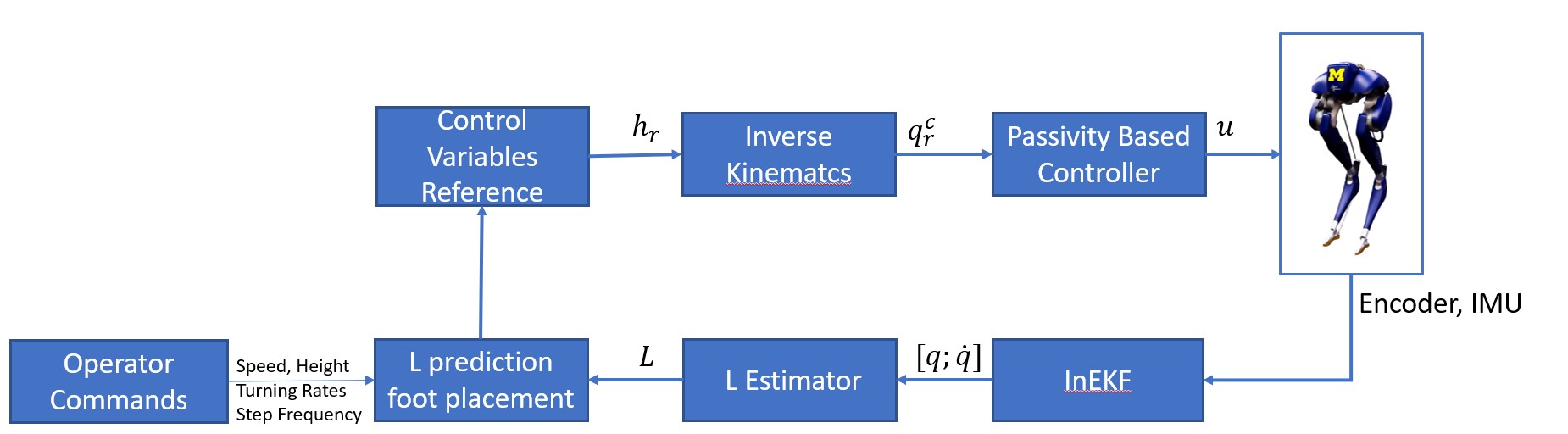}
    \caption{Block diagram of the implemented controller.}
    \label{fig:Control_Diagram}
\end{figure*}
This section discusses several issues that prevent the basic controller from being implemented on Cassie Blue. Similar issues may or may not arise on your robot.

\subsection{IMU and EKF}
In a real robot, an IMU and an EKF are needed to estimate the linear position and rotation matrix at a fixed point on the robot, along with their derivatives. Cassie uses a VectorNav IMU. We used the Contact-aided Invariant EKF developed in \cite{hartley2020contact,invariant-ekf} to estimate the torso velocity. With these signals in hand, we could estimate angular momentum about the contact point.

\subsection{Filter for Angular Momentum}
Angular Momentum about the contact toe could be estimated directly from the sensors on the robot, but it was noisy. We designed a Kalman Filter to improve the estimation. The models we used are
\begin{equation}
    \begin{aligned}
   \text{Prediction:    } & L^y(k+1) = L^y(k) + mgp_{\rm st}(k)\Delta T + \delta(k)\\
   \\
   \text{Correction:  }& L^y(k) = L^y_{\rm \;obs}(k) + \epsilon(k)
   \end{aligned}
\end{equation}

\subsection{Inverse Kinematics}
Input-Output Linearization does not work well in experiments\cite{Yukai2018,CLFKoushil,WEBUGR04}. To use a passivity-based controller for tracking that is inspired by \cite{sadeghian2017passivity}, we need to convert the reference trajectories for the variables in \eqref{eq:control variables} to reference trajectories for Cassie's actuated joints,
\begin{equation}
    q^{\rm act} = 
    \begin{bmatrix}
    \rm torso \; pitch \\
    \rm torso \; roll \\
    \rm stance \; hip \; yaw \\
    \rm swing \; hip \; yaw \\
    \rm stance \; knee \; pitch \\
    \rm swing \; hip \; roll \\
    \rm swing \; hip \; pitch \\
    \rm swing \; knee \; pitch \\
    \rm swing \; toe \; pitch \\
    \end{bmatrix}.
\end{equation}
Iterative inverse kinematics is used to convert the controlled variables in \eqref{eq:control variables} to the actuated joints. 

\textbf{Why did we not directly use actuated joints?} We used \eqref{eq:control variables} to express the controller objectives because it allows the most intuitive implementation of the basic controller on a physical robot. Also, due to our heritage of using virtual constraints, this is natural to us. Your philosophy may vary.

\subsection{Passivity-based Controller}

To arrive at a PD controller with feed forward terms, we modified the passivity based controller developed in \cite{sadeghian2017passivity}. The key modifications were:
\begin{itemize}
    \item transitioning the method to a floating base model of the robot; and 
    \item using the phase variable \eqref{eq:phase_variable}.
\end{itemize}
This provided improved tracking performance over the straight-up PD implementation in \cite{Yukai2018}.

% Though PD control on Motor is good enough for walking in place, it will create a large error when walking fast,(10 cm error in foot placement), which is not acceptable. Thus we introduce PBC to improve tracking performance to achieve fast walking.

% The floating base of Cassie in single support is:
% \begin{equation}
%     \begin{bmatrix}
%     D(q) & -J^T(q) \\
%     J(q) & 0 \\
%     \end{bmatrix}
%     \begin{bmatrix}
%     \ddot{q} \\
%     \lambda
%     \end{bmatrix}
%     + 
%     \begin{bmatrix}
%     [C(q,\dot{q})\dot{q} + G(q)] \\
%     \dot{J}(q,\dot{q})\dot{q} \\
%     \end{bmatrix}
%     =
%     \begin{bmatrix}
%     B \\
%     0
%     \end{bmatrix}
%     u
% \end{equation}
% where J is holonomic constraint, including 5 constraints provided by blade-shape foot, and 4 spring constraints, which is obtained by assuming springs are rigid. 
% By some Calculation, we could get
% \begin{equation}
%     \bar{D}(q)\ddot{q}^c + \bar{C}(q,\dot{q})\dot{q} + \bar{G}(q) = \bar{B}u
% \end{equation}

% The torque u is obtained by
% \begin{equation}
% u = \bar{B}^{-1}(\bar{D}\ddot{q}^c_r + \bar{C})\dot{q} + \bar{G}) - (\bar{C}+K_d\dot{q}^c_e) - K_pq^c_e
% \end{equation}
% Where $q^c_r$ is reference trajectory and $q^c_e$ is tracking error.
% In experiment all Coriolis term are omitted for computation efficiency.
% \yg{need to check equation later}

\subsection{Springs}
On the swing leg, the spring deflection is small and thus we are able to assume the leg to be rigid. On the stance leg, the spring deflection is not negligible. An offset is added on the knee motor to compensate this spring deflection. While there are encoders to measure the spring deflection, direct use of this leads to oscillations. The deflection of the spring is instead calculated through a simplified model.  

\subsection{COM Velocity in the Vertical Direction}

When Cassie is walking over one meter per second, the assumption that $v^z_{\rm CoM} = 0$ breaks down and \eqref{eq:AM_impact_relation_1} is no longer valid. 
Hence, we use
\begin{equation} \label{eq:AM_impact_relation_2} \small
    L^y(T_k^+) = L^y(T_k^-) + v^z_{\rm CoM}(T_k^-)(p^x_{\rm sw \to CoM}(T_k^-) - p^x_{\rm st \to CoM}(T_k^-)).
\end{equation}
From this, the foot placement is updated to
% \begin{multline} \label{eq:foot_placement_2}
%     p^{x \; \rm des}_{\rm sw \to CoM}(T_k^-) = \\
%     \frac{1}{m(H\ell \sinh(\ell T) - v^z_{\rm CoM})\cosh(\ell T)}\dot \\
%   (L^{y \; \rm des}(T_{k+1}^-) \\
%   - (L^y(T_k^-)+mv^z_{\rm CoM}(T_k^-)p^x_{\rm st \to CoM}(T_k^-)\cosh(\ell T)))
% \end{multline}
\begin{multline} \label{eq:foot_placement_2} \small
    p^{x \; \rm des}_{\rm sw \to CoM}(T_k^-) =  \frac{ L^{y \; \rm des}(T_{k+1}^-) -}{ m(H\ell \sinh(\ell T) - v^z_{\rm CoM})\cosh(\ell T) } - \\
    \frac {(L^y(T_k^-)+mv^z_{\rm CoM}(T_k^-)p^x_{\rm st \to CoM}(T_k^-))\cosh(\ell T) }{m(H\ell \sinh(\ell T) - v^z_{\rm CoM})\cosh(\ell T)}
\end{multline}

% Though \eqref{eq:AM_impact_relation_2} does not need $v^z_{\rm CoM}$ be 0, it still assumes the ground is flat. As a result, when walking uphills or downhills, it will give larger estimation errors, which in turn make the gait "wider" or "narrower", similar to the phenomenon mentioned above

% Above We discussed the effect of $v_{\rm z \; CoM}$ \textbf{at impact}. During continuous phase, the effect of $v_{\rm z \; CoM}$ and $L_{COM}$ is similar. That they affect Horizontal Linear Momentum a lot, but affect $L$ little (under the condition that the controller is \textbf{trying} to make $v^z_{\rm CoM}$ 0, it will affects more when $a^z_{\rm CoM}$ is large). Thus we ignore its effect during continuous phase.

% Though $v^z_{\rm CoM}$ is treated as a nuisance here, it could be an option to utilize to stabilize gaits, other than foot placement and step frequency. 

\section{Experimental Results} \label{Sec:ExperimentResults}

The overall controller summarized in Fig.~\ref{fig:Control_Diagram} was implemented on Cassie Blue. The closed-loop system consisting of robot and controller was evaluated in  a number of situations that are itemized below.

\begin{itemize}
    \item \textbf{Walking in a straight line on flat ground.} Cassie could walk in place and walk stably for speeds ranging from zero to $2.1$ m/s. %These results are illustrated in Fig.~\ref{} and \cite{video} at time $t=$.
    
    \item \textbf{Diagonal Walking.} Cassie is able to walk simultaneously forward and sideways on grass, at roughly $1$ m/s in each direction.% These results are illustrated in Fig.~\ref{} and \cite{video} at time $t=$.
    
    \item \textbf{Sharp turn.} While walking at roughly $1$ m/s, Cassie Blue effected a $90^o$ turn, without slowing down. %These results are illustrated in Fig.~\ref{} and \cite{video} at time $t=$.
    
    \item \textbf{Rejecting the classical kick to the base of the hips.} Cassie was able to remain upright under ``moderate'' kicks in the longitudinal direction. The disturbance rejection in the lateral direction is not as robust as the longitudinal, which is mainly caused by Cassie's physical design: small hip roll motor position limits. %These results are illustrated in Fig.~\ref{} and \cite{video} at time $t=$.
    
    \item \textbf{Finally we address walking on rough ground}. Cassie Blue was tested on the iconic Wave Field of the University of Michigan North Campus. The foot clearance was increased from 10 cm to 20 cm to handle the highly undulating terrain. Cassie is able to walk through the``valley'' between the large humps with ease at a walking pace of roughly $0.75$ m/s, without falling in all tests. The row of ridges running east to west in the Wave Field are roughly 60 cm high, with a sinusoidal structure. We estimate the maximum slope to be 40 degrees. Cassie is able to cross several of the large humps in a row, but also fell multiple times. On a more gentle, straight grassy slope of roughly 22 degrees near the laboratory, Cassie can walk up it with no difficulty whatsoever. %These results are illustrated in Fig.~\ref{} and \cite{video} at time $t=$.
\end{itemize} 

\begin{figure}
\begin{subfigure}{.242\textwidth}
  \centering
  \includegraphics[width=\textwidth]{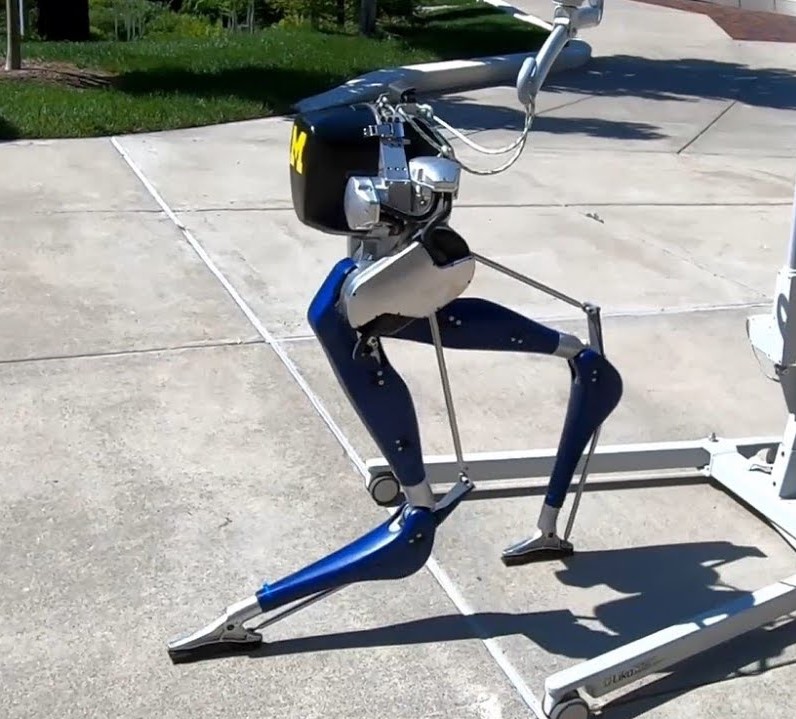}
  \caption{Fast Walking}
  \label{fig:FastWalking}
\end{subfigure}
\begin{subfigure}{.238\textwidth}
  \centering
  \includegraphics[width=\textwidth]{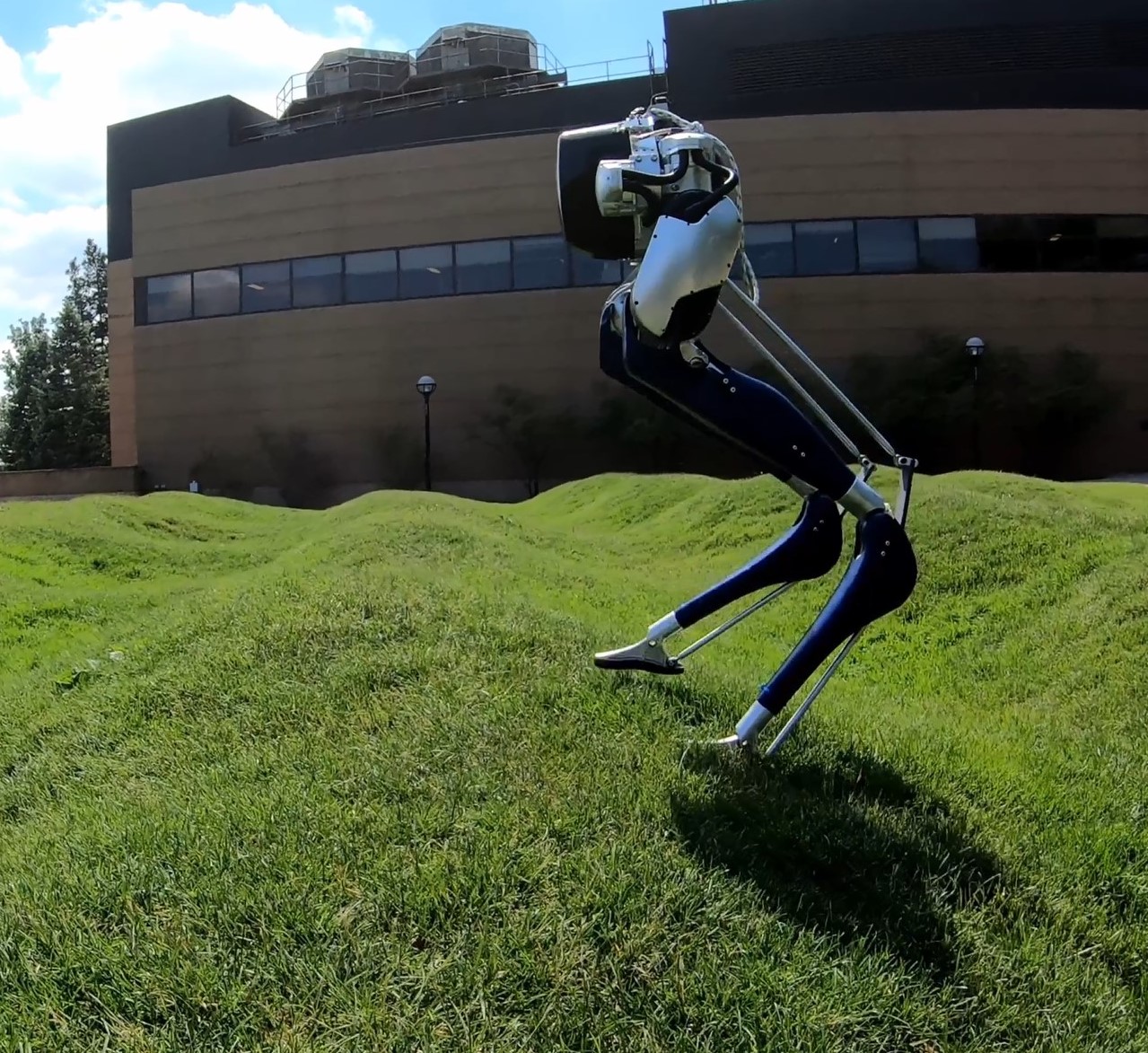}
  \caption{Rough Terrain}
  \label{fig:RoughTerrain}
\end{subfigure}
\begin{subfigure}{.488\textwidth}
  \centering
  \includegraphics[width=\textwidth]{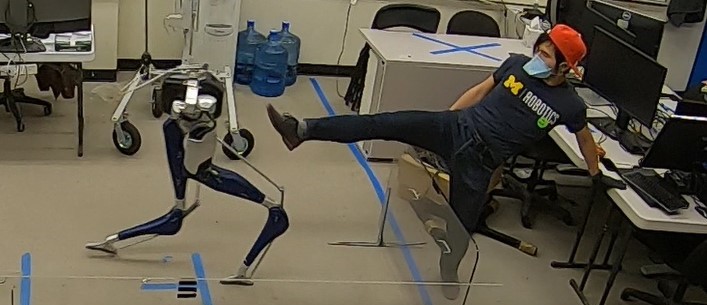}
  \caption{Disturbance Rejection}
  \label{fig:KickTest}
\end{subfigure}
\begin{subfigure}{.488\textwidth}
  \centering
  \includegraphics[width=\textwidth]{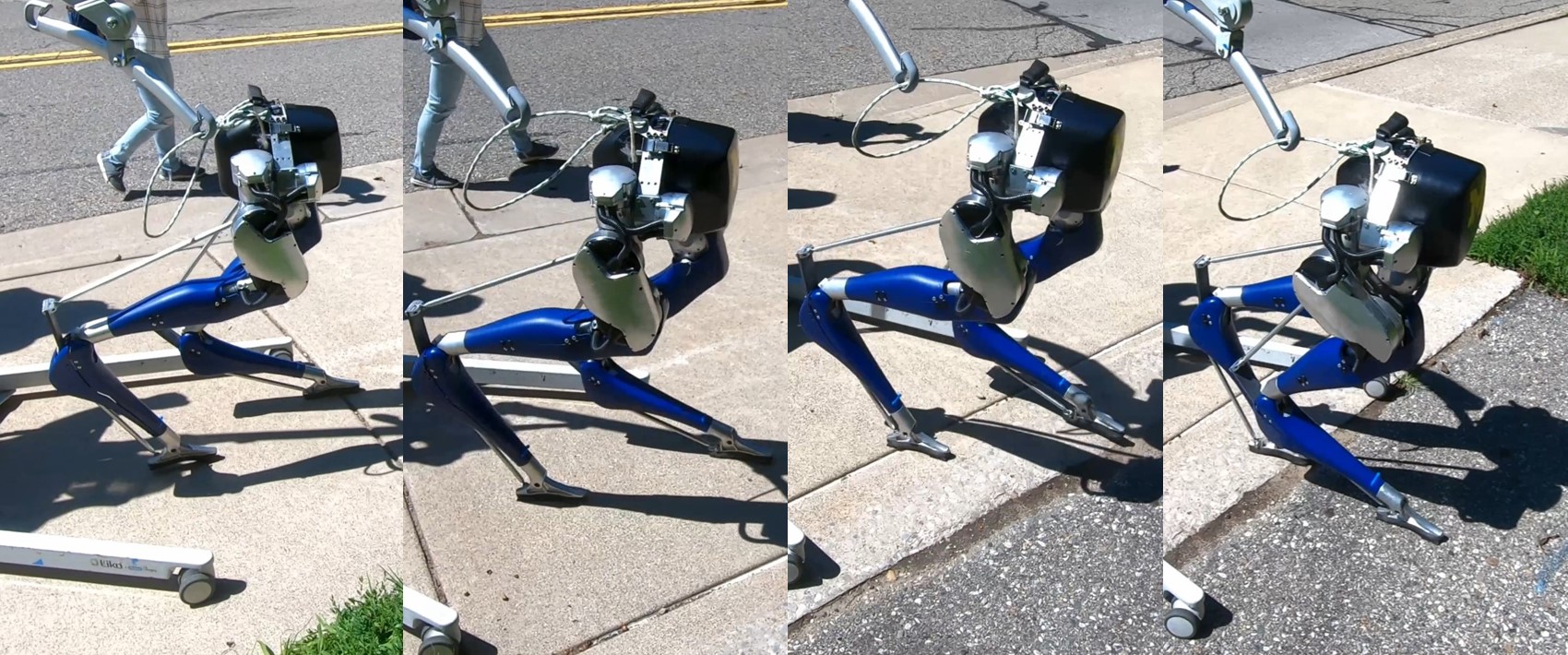}
  \caption{A Fast 90 Degree Turn with a Long Stride}
  \label{fig:Turning}
\end{subfigure}
\caption{Images from several closed-loop experiments conducted with Cassie Blue and the controller developed in this paper.Short Footage of those experiments are compiled in video \cite{AMLIP_Video}. Longer versions can be found in \cite{LabYoutube}}
\label{fig:ExperimentResult}
\end{figure}

\section{Conclusions} \label{Sec:Conclusion}
We argued that angular momentum about the contact point is a superior variable for planning step placement in a LIP-based controller. We believe the same will hold on many other control strategies. This paper limited itself to a LIP-style controller so that other locomotion groups could clearly assess the benefits of angular momentum about the contact point. 

Using our new controller, Cassie was able to accomplish a wide range of tasks with nothing more than common sense task-based tuning: a higher step frequency to walk at 2.1 m/s and extra foot clearance to walk over slopes exceeding 15 degrees. Moreover, in the current implementation, there is no optimization of trajectories used in the implementation on Cassie. The robot's performance is currently limited by the hand-designed trajectories leading to joint-limit violations and foot slippage. These limitations will be alleviated by incorporating optimization.

The current controller tries its best to maintain a zero center of mass velocity in the $z$-direction. This simplifies the transition formula for the angular momentum at impact. Our next publication will explain how to exploit changes in the vertical component of the center of mass velocity in order to better achieve a desired angular momentum: foot placement plus $v^{z}_{\rm CoM}$! 

%  \FloatBarrier
% \clearpage
% \frontmatter
\bibliographystyle{plain}
\balance
\bibliography{BibFiles/Bib2020July.bib}
% \bibliography{BibFiles/bib2.bib}

\end{document}